\documentclass[11pt]{article}

\usepackage[final]{coling}

\usepackage{times}
\usepackage{latexsym}
\usepackage{enumitem}
\usepackage{amsmath}
\usepackage{booktabs}
\usepackage{array} 
\usepackage{float}
\usepackage{multirow}
\usepackage[utf8]{inputenc}
\usepackage{CJKutf8} 
\usepackage{fontspec}
\usepackage[T1]{fontenc}

\usepackage[utf8]{inputenc}

\usepackage{microtype}

\usepackage{inconsolata}

\usepackage{graphicx}

\title{CUTE: A Multilingual Dataset for Enhancing Cross-Lingual Knowledge Transfer in Low-Resource Languages}

\author{
  \textbf{Wenhao Zhuang\textsuperscript{1,2}},
  \textbf{Yuan Sun\textsuperscript{1,2,*}}\\
  \textsuperscript{1}Minzu University of China, Beijing, China\\
  \textsuperscript{2}National Language Resource Monitoring \& Research Center Minority Languages Branch\\
  Emails: \texttt{sdrz\_zwh@163.com}, 
  \texttt{sunyuan@muc.edu.cn}\\
  {* Corresponding author: Yuan Sun}
}

\begin{document}
\maketitle
\begin{abstract}

Large Language Models (LLMs) demonstrate exceptional zero-shot capabilities in various NLP tasks, significantly enhancing user experience and efficiency. However, this advantage is primarily limited to resource-rich languages. For the diverse array of low-resource languages, support remains inadequate, with the scarcity of training corpora considered the primary cause. We construct and open-source CUTE (\textbf{C}hinese, \textbf{U}yghur, \textbf{T}ibetan, \textbf{E}nglish) dataset, consisting of two 25GB sets of four-language corpora (one parallel and one non-parallel), obtained through machine translation. CUTE encompasses two resource-rich languages (Chinese and English) and two low-resource languages (Uyghur and Tibetan). Prior to constructing CUTE, human assessment validates that the machine translation quality between Chinese-Uyghur and Chinese-Tibetan approaches that of Chinese-English translation. CUTE represents the largest open-source corpus for Uyghur and Tibetan languages to date, and we demonstrate its effectiveness in enhancing LLMs' ability to process low-resource languages while investigating the role of corpus parallelism in cross-lingual transfer learning. The CUTE corpus and related models are made publicly available to the research community\footnote{\url{https://github.com/CMLI-NLP/CUTE}}.

\end{abstract}

\section{Introduction}

The current LLMs demonstrate remarkable capabilities in resource-rich languages. However, their performance is limited for numerous resource-poor languages~\cite{Ebrahimi2021AmericasNLIEZ,chowdhery2023palm}. Even powerful multilingual models such as XLM-R~\cite{Conneau2019UnsupervisedCR}, mT5~\cite{Xue2020mT5AM}, and NLLB~\cite{costa2022no} support only approximately 100-200 languages, leaving nearly 7,000 low-resource languages untapped~\cite{Esch2022WritingSA}. Among these are several low-resource languages with significant numbers of speakers. Uyghur and Tibetan, two low-resource minority languages in China, have over 13 million and 8 million speakers respectively. However, known LLMs have yet to achieve adequate support for these two languages.

Existing multilingual datasets, such as OSCAR~\cite{Abadji2022TowardsAC} and CulturaX~\cite{nguyen2024culturax}, include Uyghur and Tibetan languages. The CC100 dataset~\cite{Conneau2019UnsupervisedCR}, used for training XLM-R, also contains a small amount of Uyghur text. However, these datasets still exhibit several limitations, including insufficient data volume, misidentification of languages, and imbalanced data distribution~\cite{zhang2024mc}. MC$^\text{2}$ represents the largest open-source multilingual corpus of Chinese ethnic minority languages to date. It comprises crawled, integrated, and cleaned data from existing minority language sources, including Uyghur and Tibetan~\cite{zhang2024mc}. Nevertheless, the scale of this dataset remains relatively small, not exceeding 3GB in size.

To enhance LLMs' ability to process low-resource languages, continued pre-training or adding adapters are common approaches~\cite{Yong2022BLOOM1AL,zhang2024mc,cahyawijaya2023instructalign,Jin2022DatalessKF}. However, continued pre-training typically requires substantial unlabeled text for learning language representations, while question-answer pairs in low-resource languages for fine-tuning are even more challenging to obtain. Existing low-resource corpora are often insufficient to effectively update LLM parameters~\cite{cahyawijaya2024llms}. To rapidly address the scale issue of low-resource corpora, one solution involves using machine translation models to translate training corpora from resource-rich languages like English and Chinese into low-resource languages. However, this approach raises two primary concerns: (1) the accuracy of the translation process may not be guaranteed~\cite{Ebing2023ToTO}. (2) cultural nuances inherent in the languages may be lost or erroneously propagated during translation~\cite{zhang2024mc,Liu2023AreML}.

LLMs demonstrate strong comprehension and instruction-following capabilities across multiple high-resource languages. This multilingual proficiency largely relies on cross-lingual sentence embeddings. Specifically, cross-lingual sentence embeddings encode multilingual text into a unified semantic representation space, where sentences with similar meanings in different languages are mapped to proximate vector locations~\cite{Conneau2019UnsupervisedCR,Devlin2019BERTPO,Lample2019CrosslingualLM}. However, significant representational disparities exist between cross-lingual word representations of low-resource languages and those of high-resource languages in current LLMs~\cite{Miao2024EnhancingCS}. Given the word-level representational differences, sentence-level cross-lingual representation alignment faces even more severe challenges. Achieving semantic representation alignment between low-resource and high-resource languages would provide more opportunities for transferring knowledge from high-resource languages to low-resource languages.Parallel corpora play a crucial role as bridges in transfer learning~\cite{Pham2024UniBridgeAU}. However, accurately assessing their impact on LLMs and determining whether parallel corpora can facilitate more effective cross-lingual knowledge transfer requires a parallel corpus of sufficient scale and quality.

To advance research and development of Uyghur and Tibetan in LLMs, validate the reliability of machine translation in generating low-resource data, and explore the impact of parallel corpora in knowledge transfer from high-resource to low-resource languages, this paper introduces the CUTE (\textbf{C}hinese, \textbf{U}yghur, \textbf{T}ibetan, \textbf{E}nglish) dataset. It comprises two equal-sized four-language corpora: one parallel in content and the other non-parallel, totaling approximately 50GB. The Uyghur and Tibetan components are ten times larger than the current largest open-source MC$^\text{2}$ dataset.

The CUTE dataset addresses the common issues present in the aforementioned datasets that include Uyghur and Tibetan languages. Notably, the CUTE dataset offers several improvements. First, it boasts a significantly larger scale. Additionally, the use of machine translation in CUTE eliminates the problem of language misidentification. Furthermore, the dataset features a more balanced distribution of data across languages and content domains. The mutual parallelism among the four languages in CUTE also provides expanded opportunities for research in machine translation and cross-lingual knowledge transfer.

In summary, we make the following contributions:

\begin{itemize}[topsep=0pt,itemsep=0pt]
  \item We construct and open-source the CUTE dataset, a large-scale multilingual corpus with parallel and non-parallel data in two high-resource languages (Chinese and English) and two low-resource languages (Uyghur and Tibetan), facilitating LLM training and evaluation for minority languages.
  
  \item We propose a novel approach for LLM training using parallel and non-parallel corpora through vocabulary expansion and embedding initialization. Using this method, we develop and release two versions of CUTE-Llama trained on different corpus types.
  
  \item We validate the effectiveness of parallel corpora in cross-lingual knowledge transfer through zero-shot experiments on various downstream tasks, showing that parallel data enables more effective knowledge transfer from high-resource to low-resource languages.
\end{itemize}

\section{Related Works}



\paragraph{Low-resource Language Corpora}

In recent years, several large-scale multilingual corpora have emerged to support NLP tasks for low-resource languages. Datasets such as OSCAR~\cite{Abadji2022TowardsAC}, CulturaX~\cite{nguyen2024culturax}, and MADLAD-400~\cite{Kudugunta2023MADLAD400AM} provide rich cross-lingual text for multilingual model training through web crawling and multi-source integration. The ROOTS dataset emphasizes openness and traceability, implementing strict management in data collection and cleaning for low-resource languages~\cite{laurenccon2022bigscience}. OPUS, as an open-source parallel corpus, offers extensive bilingual data for machine translation task~\cite{tiedemann-2012-parallel}s. Meta's FLORES-200 and NLLB projects focus on evaluating and improving translation performance for low-resource languages, covering 200 languages and significantly advancing cross-lingual knowledge transfer research~\cite{costa2022no}. Although these corpora demonstrate excellent performance in processing high-resource languages, they still face challenges such as insufficient data volume, language misidentification, and imbalanced content distribution for low-resource languages (e.g., Uyghur, Tibetan). Future research directions should prioritize increasing data collection for low-resource languages and improving the quality of existing corpora.


\paragraph{NLP Development for Low-Resource Languages}

NLP research for low-resource languages has made some progress in recent years but still faces numerous challenges. In terms of datasets, research primarily focuses on tasks such as text classification~\cite{Qun2017EndtoEndNT,Yang2022CINOAC,deng2023milmo}, machine reading comprehension~\cite{sun2021construction}, instruction following~\cite{zhuang2024tifd}, and machine translation~\cite{Zhang2024TeachingLL}. However, these datasets are typically limited in scale and cover a narrow range of languages. Regarding models, pre-trained models specifically developed for Chinese minority languages, such as CINO~\cite{Yang2022CINOAC}, MiLMo~\cite{deng2023milmo}, CMPT~\cite{Li2022AMA}, and TiLamb~\cite{zhuang2024tilamb}, have achieved certain breakthroughs in processing languages like Uyghur and Tibetan through techniques including multilingual pre-training and vocabulary expansion. Nevertheless, the pre-training corpora for these models are generally not publicly available, which limits the reproducibility and further development of research.
Existing LLMs contribute minimally to improving the processing capabilities for low-resource languages, primarily due to the lack of high-quality instruction data. Knowledge distillation methods from teacher models prove ineffective for these languages, as current LLMs already perform inadequately in low-resource languages such as Uyghur and Tibetan. Moreover, finding suitable annotators capable of writing high-quality instruction samples is challenging due to the high requirements for creative thinking and professional expertise~\cite{Li2024XInstructionAL}. These factors collectively constrain the development of NLP for low-resource languages in China, highlighting urgent research needs in data collection, model optimization, and cross-lingual knowledge transfer.

\paragraph{Cross-lingual Knowledge Transfer}

Optimizing cross-lingual knowledge transfer is a key strategy for addressing NLP challenges in low-resource languages. Multilingual pre-trained models such as XLM-R~\cite{Conneau2019UnsupervisedCR} and mBERT~\cite{Devlin2019BERTPO} currently serve as powerful tools for effective cross-lingual transfer, yet they still face challenges when applied to low-resource languages. In recent years, ICL (In-Context Learning)~\cite{brown2020language} and few shot learning have shown potential to adapt large language models to new tasks. However, their effectiveness in helping models understand under-trained low-resource languages remains limited. To further enhance the efficacy of cross-lingual knowledge transfer, this study focuses on two critical directions: evaluating the effectiveness of machine-translated low-resource language data in model training, and investigating the impact of parallel data on cross-lingual knowledge transfer.



\section{CUTE Dataset and CUTE-Llama}

This section provides a detailed description of the scale of the CUTE dataset, which encompasses Chinese, Uyghur, Tibetan, and English. It represents the largest open-source dataset for Uyghur and Tibetan languages in China to date. It is important to note that Chinese and English, as resource-rich languages, are well-supported in the majority of LLMs. In contrast, Uyghur and Tibetan, as low-resource languages, often exhibit suboptimal performance in nearly all LLMs.

Additionally, this section elucidates the construction and training process of CUTE-Llama, as well as the model's evolving adaptability to Uyghur and Tibetan languages during the training phase.

\subsection{Machine Translation for Chinese Minority Languages}
Low-resource languages exhibit a significant disparity in data acquisition methods and scale compared to resource-rich languages. One viable approach to build large datasets for low-resource languages is using machine translation. This method converts existing training corpora from resource-rich languages into low-resource language data. However, ensuring translation quality becomes a critical consideration in this process.

The combined number of Uyghur and Tibetan language users in China exceeds 20 million. Neural machine translation technologies between various minority languages, with Chinese as the pivot, have developed over an extended period and have now reached maturity. Given that bidirectional translation between Chinese and English, both resource-rich languages, has attained a considerable level of reliability, we have established scoring criteria to evaluate the quality of machine translation from Chinese to Uyghur and Tibetan. Native speakers of these respective minority languages have been invited to conduct the assessment.

\paragraph{Evaluation Criteria}

We have established unified translation standards with Chinese as the source language and Uyghur, Tibetan, and English as the target languages, as shown in Table~\ref{tab:evaluation_criteria}.
\begin{table*}[ht]
\small
\centering
\setlength\tabcolsep{8pt}  
\renewcommand{\arraystretch}{1.3}  
\begin{tabular}{>{\centering\arraybackslash}m{0.08\textwidth} p{0.82\textwidth}}  
\toprule
\textbf{Score} & \multicolumn{1}{c}{\textbf{Description}} \\  
\midrule
10 & Perfect translation, accurate and fluent, fully consistent with the original style and meaning. \\
\addlinespace[0.5em]
8--9 & Generally accurate, natural and fluent, with only minor errors or improprieties. \\
\addlinespace[0.5em]
6--7 & Overall meaning is understandable, but with noticeable errors or awkward phrasing that affect partial comprehension. \\
\addlinespace[0.5em]
4--5 & Partially understandable, but with serious errors that impact overall comprehension. \\
\addlinespace[0.5em]
2--3 & Mostly incomprehensible, with only a few words correctly translated. \\
\addlinespace[0.5em]
0--1 & Completely incomprehensible or unrelated to the original text. \\
\bottomrule
\end{tabular}
\caption{Machine Translation Quality Evaluation Criteria: Universal Scoring Standards for Chinese-Uyghur, Chinese-Tibetan, and Chinese-English Translation Directions in the CUTE Dataset}
\label{tab:evaluation_criteria}
\end{table*}




\paragraph{Human Evaluation}

We randomly select 500 Chinese texts from the SkyPile-150B dataset~\cite{wei2023skywork} and generate corresponding parallel sentence pairs in Uyghur, Tibetan, and English through machine translation, with 500 pairs for each language. For Chinese-English translation, we employ the Google Translate system, while specialized machine translation models are used for Chinese-Uyghur and Chinese-Tibetan translations.

\begin{figure}[ht]
\centering
\includegraphics[scale=0.35]{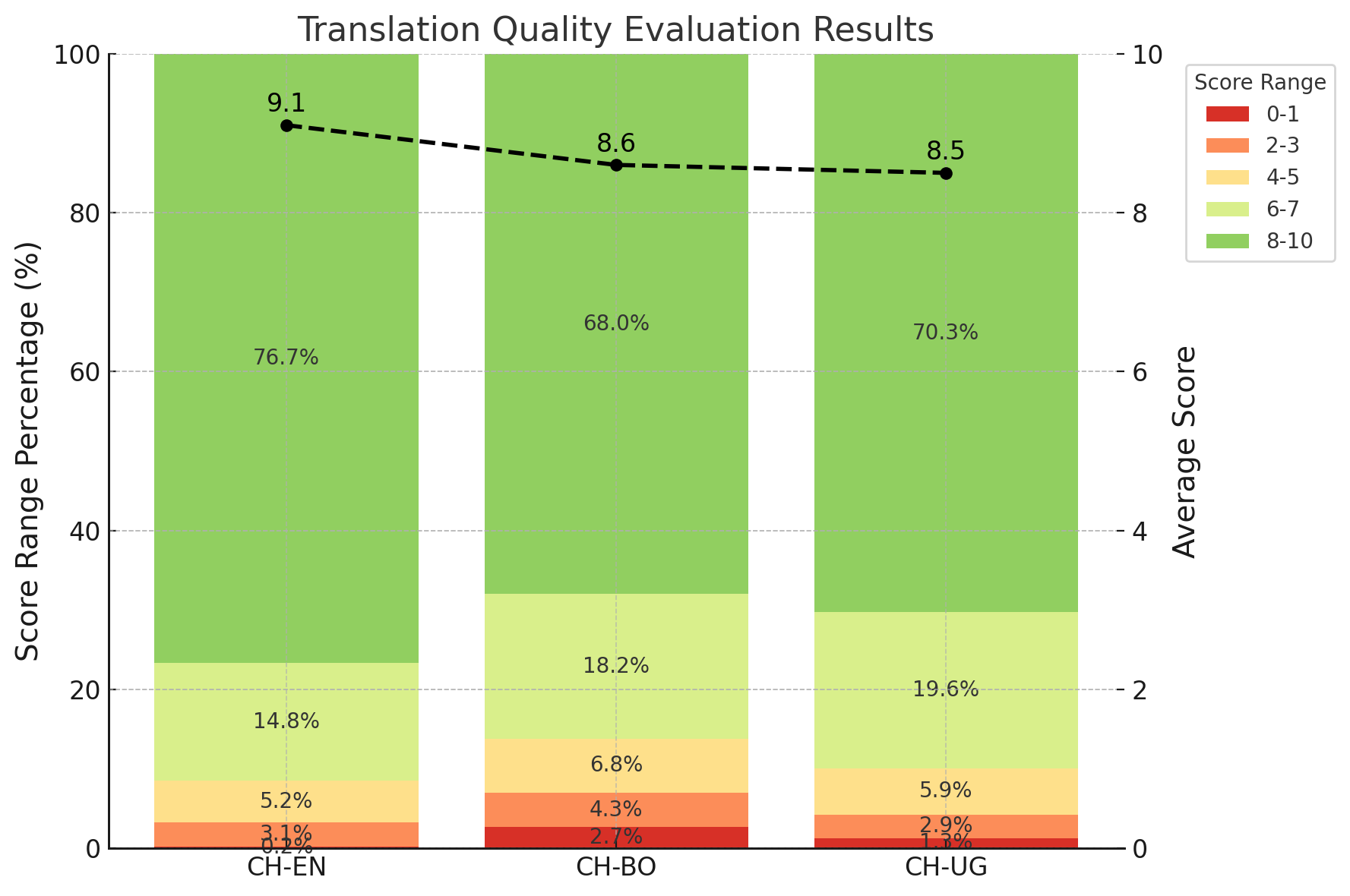}
\caption{Human evaluation results of Chinese to English, Tibetan, and Uyghur translations. The stacked bar chart displays the distribution of translation quality scores across five score ranges (0-1, 2-3, 4-5, 6-7, 8-10) for each language pair, while the black dashed line represents the average scores.}
\label{fig:evaluation_criteria}
\end{figure}

To evaluate the translation quality, we invite three Uyghur and three Tibetan graduate students as evaluators. These evaluators are native speakers of their respective ethnicities and are also proficient in Chinese. For the assessment of Chinese-English translations, we recruit three graduate students with excellent English communication skills. After standardizing the translation scoring criteria, they independently complete their respective translation scoring tasks. The final human evaluation results for Chinese translations into Uyghur, Tibetan, and English are shown in Figure~\ref{fig:evaluation_criteria}.

\paragraph{Results Analysis}
The Chinese-English (CH-EN) translation performs best, achieving an average score of 9.1, with 76.7\% of translations falling within the 8-10 score range, demonstrating high accuracy and fluency. Notably, the quality of Chinese-Uyghur (CH-UG) and Chinese-Tibetan (CH-BO) translations closely approaches that of Chinese-English, with average scores of 8.5 and 8.6 respectively, also attaining the level of "generally accurate and naturally fluent." All three language pairs exhibit a pronounced right-skewed distribution, with over 90\% of translations scoring above 6, indicating that the vast majority of translations accurately convey the overall meaning of the original text. These results highlight the balanced high-level performance of current machine translation systems in handling translations between Chinese and English, Uyghur, and Tibetan. In particular, the quality of Chinese-Uyghur and Chinese-Tibetan translations now approaches that of Chinese-English translation.

\subsection{CUTE Dataset Language Distribution}
The CUTE dataset utilizes machine translation to translate a small portion of the SkyPile-150B dataset into Uyghur, Tibetan, and English. SkyPile-150B is a dataset specifically designed for pre-training large-scale Chinese language models, containing approximately 150 billion tokens primarily sourced from a wide range of Chinese internet web content. This dataset undergoes rigorous deduplication and sensitive information filtering to ensure data quality and safety.

CUTE comprises two sets of corpora, each containing four languages. The first set consists of parallel corpora in four languages, achieving a 99.98\% similarity in content parallelism. The second set includes non-parallel corpora in four languages, with the English portion identical to the first set, while the remaining three languages differ. The specific scale of CUTE is presented in Table~\ref{tab:cute_dataset_distribution}.

\begin{table}[H]
\small
\centering
\setlength{\tabcolsep}{10pt}
\begin{tabular}{@{}lrrrr@{}}
\toprule
& \multicolumn{2}{c}{\textbf{CUTE-P}} & \multicolumn{2}{c}{\textbf{CUTE-NP}} \\
\cmidrule(lr){2-3} \cmidrule(l){4-5}
\textbf{Lang.} & \textbf{Lines} & \textbf{Size} & \textbf{Lines} & \textbf{Size} \\
\midrule
ZH & 933,946 & 2.62 & 1,000,609 & 2.64 \\
EN & 933,989 & 3.49 & 933,989 & 3.49 \\
UG & 934,002 & 7.37 & 1,010,381 & 7.77 \\
BO & 934,140 & 11.22 & 989,723 & 11.90 \\
\midrule
\textbf{Total} & 3,736,077 & 24.70 & 3,934,702 & 25.80 \\
\bottomrule
\end{tabular}
\caption{Distribution of CUTE dataset. CUTE-P: parallel corpus, CUTE-NP: non-parallel corpus. Language codes: ZH (Chinese), EN (English), UG (Uyghur), BO (Tibetan). Size in GB.}
\label{tab:cute_dataset_distribution}
\end{table}

\subsection{Analysis of Document Length Distribution}

\begin{figure}[H]
\centering
\includegraphics[scale=0.3]{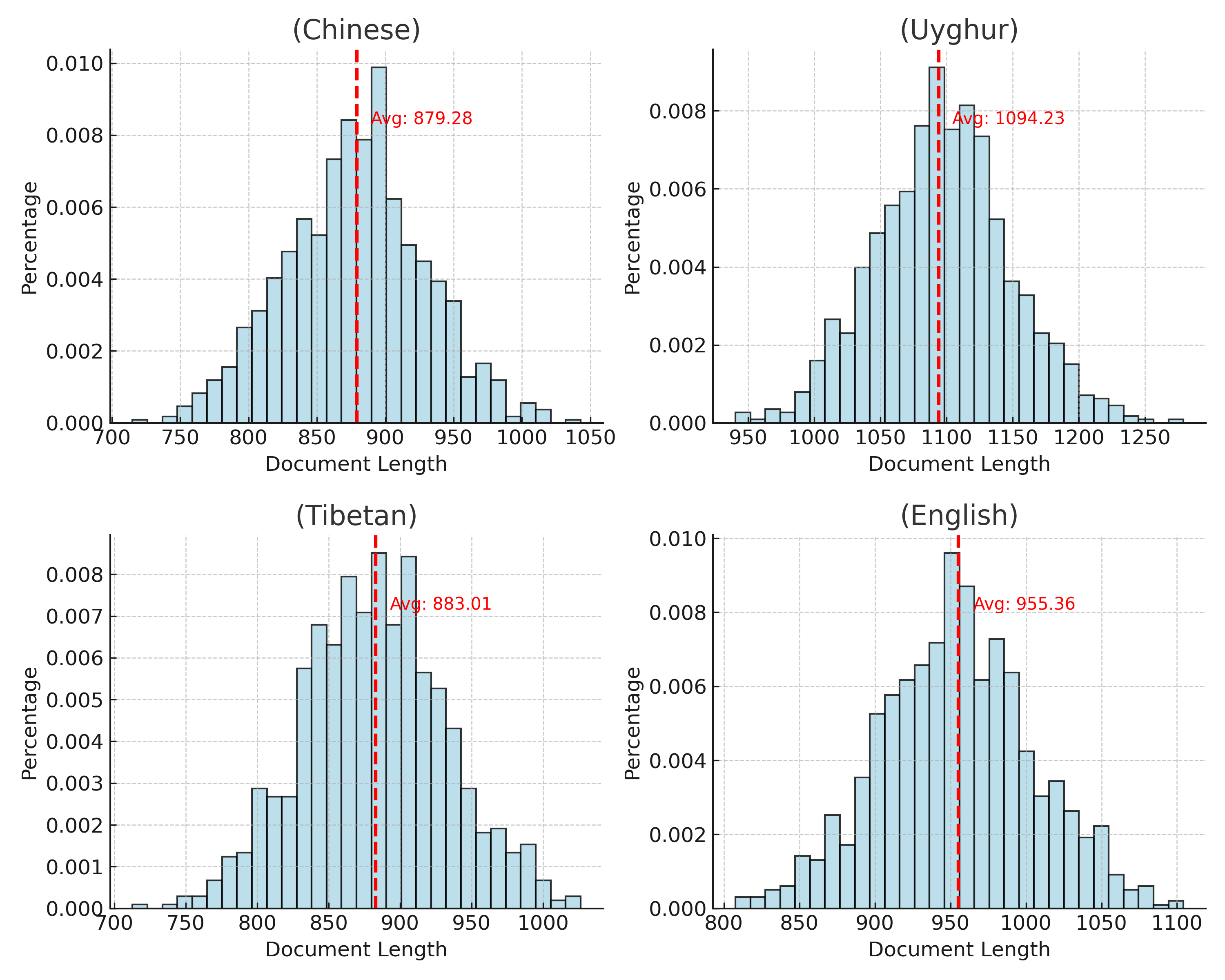}
\caption{Document length distribution across Chinese, Uyghur, Tibetan, and English in the CUTE dataset, based on token counts calculated using the CUTE-Llama tokenizer. The red dashed lines represent the average document length for each language.}
\label{fig:doc_length_distribution}
\end{figure}

Figure ~\ref{fig:doc_length_distribution} illustrates the document length distribution for four languages (Chinese, Uyghur, Tibetan, and English) in the CUTE dataset. By analyzing the document lengths for each language, we observe significant variations in the average document length across languages. Uyghur exhibits the highest average document length at 1,094.23 tokens, substantially exceeding the other three languages. In contrast, English, Tibetan, and Chinese demonstrate relatively similar average document lengths of 955.36, 883.01, and 879.28 tokens, respectively.

\subsection{CUTE-Llama Vocabulary Training}

CUTE-Llama is based on the Llama2 model architecture~\cite{touvron2023llama2}. The original tokenization model of Llama2 is trained using the SentencePiece~\cite{kudo2018sentencepiece} library, employing the Byte Pair Encoding (BPE) algorithm. This algorithm constructs the vocabulary by merging common byte pairs, enabling effective processing of various languages. Considering that Llama2 is well-adapted for English but lacks sufficient support for Chinese, Uyghur, and Tibetan, we train separate vocabularies of 6,000 tokens each for these three languages. The training content, while distinct from the CUTE dataset, is of comparable scale. These newly trained vocabularies are subsequently merged with the Llama2 vocabulary. The parameters for vocabulary training are presented in Table \ref{tab:vocabulary_training_parameters}.

\begin{table}[H]
\centering
\setlength{\tabcolsep}{20pt}
\begin{tabular}{lr}
\toprule
\textbf{Parameter} & \textbf{Value} \\
\midrule
vocab\_size & 6,000 \\
model\_type & bpe \\
split\_digits & True \\
max\_sentence\_length & 10,000 \\
byte\_fallback & True \\
\bottomrule
\end{tabular}
\caption{SentencePiece training parameters for Chinese, Tibetan, and Uyghur vocabularies. Vocabulary sizes after merging with the original Llama model: 36,820 (Chinese), 42,353 (Tibetan), and 47,905 (Uyghur) tokens, from an initial 32,000.}
\label{tab:vocabulary_training_parameters}
\end{table}

\subsection{CUTE-Llama Training}
We trained the CUTE-Llama model using key hyperparameters as shown in Table \ref{tab:cute_llama_training}. The training process was conducted on 8 NVIDIA H800 GPUs for approximately 18 hours to obtain one CUTE-Llama model.

\begin{table}[t]
\centering
\setlength{\tabcolsep}{20pt}
\begin{tabular}{lr}
\toprule
\textbf{Hyperparameter} & \textbf{Value} \\
\midrule
Max Sequence Length & 4,096 \\
Batch Size & 256 \\
Learning Rate & 1e-4 \\
Warmup Steps & 100 \\
Epoch & 1 \\
Data Type & BF16 \\
\bottomrule
\end{tabular}
\caption{Key hyperparameters for CUTE-Llama training. }
\label{tab:cute_llama_training}
\end{table}

\subsection{Perplexity Analysis Across Training Stages}

To evaluate the model's performance across different languages throughout the training process, we conducted a perplexity (PPL) analysis using 1,000 samples for each language that were not included in the training set. Table~\ref{tab:ppl_analysis} presents the perplexity scores at various stages of model development.

\begin{table}[t]
\centering
\small
\setlength{\tabcolsep}{4pt}
\begin{tabular}{lrrrr}
\toprule
\textbf{Stage} & \textbf{Tibetan} & \textbf{Uyghur} & \textbf{Chinese} & \textbf{English} \\
\midrule
Original & 3.24 & 5.24 & 5.23 & 7.24 \\
Post-Exp & 50,633 & 16,317 & 822 & 7.25 \\
CUTE-P & 12.00 & 5.50 & 9.93 & 4.75 \\
CUTE-NP & 11.84 & 5.40 & 10.41 & 4.67 \\
\bottomrule
\end{tabular}
\caption{Perplexity scores across training stages. Stages: Original (initial Llama2), Post-Exp (after vocabulary expansion), CUTE-P (after parallel corpus training), CUTE-NP (after non-parallel corpus training).}
\label{tab:ppl_analysis}
\end{table}

The results reveal an intriguing phenomenon in the original Llama2 model, where Tibetan, Uyghur, and Chinese show lower perplexity than English. This counterintuitive outcome likely stems from the model's treatment of unfamiliar scripts as sequences of unknown tokens, leading to simplified character-level predictions. The vocabulary expansion initially causes a sharp increase in perplexity for these languages, reflecting the model's adjustment to the new token distribution. Subsequent training on both parallel and non-parallel corpora significantly improves performance across all languages, resulting in a more balanced multilingual model. This process demonstrates the effectiveness of our approach in adapting Llama2 to handle Tibetan, Uyghur, and Chinese while maintaining its English capabilities.

\section{Evaluation Tasks and Results}

To evaluate the practical value of the CUTE dataset and investigate the impact of multilingual parallel corpora on cross-lingual knowledge transfer, we construct two foundation models based on the Llama2-7B architecture: CUTE-Llama-P (trained with parallel corpora) and CUTE-Llama-NP (trained with non-parallel corpora). The vocabularies of both models are expanded to include Chinese, Uyghur, and Tibetan, with the embeddings of newly added tokens initialized using mean values. Subsequently, we conduct continuous pre-training on these models using parallel and non-parallel corpora from CUTE, respectively.

Our experimental design is as follows: We first fine-tune these two foundation models using downstream task data from resource-rich languages, then assess the cross-lingual zero-shot transfer learning capabilities of the fine-tuned models on low-resource languages. This process aims to validate the efficacy of the CUTE dataset and compare the differences between parallel and non-parallel corpora in facilitating cross-lingual knowledge transfer.
\subsection{Evaluation Datasets}


Public test sets for Uyghur and Tibetan are limited in quantity and narrow in domain coverage. To address this issue, we identify corresponding datasets with similar training tasks in resource-rich languages. \textbf{WCM-v2}~\cite{Yang2022CINOAC} is a multilingual dataset containing 10 categories of text classification tasks, including classification tasks for Chinese, Uyghur, and Tibetan. As the training set of WCM-v2 only contains Chinese data, we fine-tune the model using the Chinese training set and evaluate it on test sets in Chinese, Uyghur, and Tibetan.
\textbf{TibetanQA}~\cite{sun2021construction} is a Tibetan machine reading comprehension dataset, primarily used to assess model performance on extractive reading comprehension tasks. \textbf{CMRC}~\cite{cui2019span}, as a Chinese machine reading comprehension dataset, shares the same task type as TibetanQA. Therefore, we fine-tune CUTE-Llama using CMRC and utilize TibetanQA to test the model's transfer ability from Chinese to Tibetan.
\textbf{SQuAD}~\cite{rajpurkar2016squad} is a widely used English machine reading comprehension dataset containing question-answer pairs from Wikipedia articles. We fine-tune CUTE-Llama using SQuAD and evaluate its transfer ability from English to Tibetan through TibetanQA.
The Chinese relation extraction dataset released by Baidu, which we refer to as \textbf{Baidu-KG} for convenience, encompasses type definitions for 50 relation extraction tasks. Based on this, we construct a Tibetan relation extraction dataset, which we name \textbf{Tibetan-KG}, containing 11 relation types to test the model's performance.
The \textbf{Flores-200}~\cite{costa2022no} dataset includes machine translation tasks for Tibetan and Uyghur. We employ a few-shot prompting approach to complete this task.

\subsection{Compared Models}
We compare the CUTE-Llama model with CINO~\cite{Yang2022CINOAC}, Llama2-7B~\cite{touvron2023llama2}, BLOOM7.1B~\cite{workshop2022bloom}, and Llama3.1-8~\cite{dubey2024llama}. CINO is a pre-trained model for ethnic minority languages in China, incorporating training data from Chinese, Uyghur, and Tibetan languages. BLOOM-7.1B is pre-trained on more than 45 languages, while both Llama2-7B and Llama3.1-8B are multilingual models developed by Meta.

\subsection{Experimental Setup and Results}
We compare and analyze the experimental results for text classification, relation extraction, machine reading comprehension, and translation in this section.

\paragraph{Text Classification}
We evaluate the models' performance on the WCM-v2 dataset, which includes Chinese (zh), Tibetan (bo), and Uyghur (ug) languages. The dataset comprises 32,000 Chinese samples for training. For testing, we use 4,000 Chinese, 1,110 Tibetan, and 300 Uyghur samples. Models are fine-tuned on the Chinese training data and tested on all three languages to assess zero-shot transfer capabilities. Table \ref{tab:performance_comparison} presents the classification results.

\begin{table*}[ht]
\centering
\resizebox{\textwidth}{!}{%
\setlength\tabcolsep{6pt}
\begin{tabular}{l|ccc|cc}
\toprule
\multirow{2}{*}{\textbf{Model}} & \multicolumn{3}{c|}{\textbf{Classification (Accuracy / F1)}} & \multicolumn{2}{c}{\textbf{Average}} \\
& \texttt{zh} & \texttt{bo} & \texttt{ug} & Minorities & All \\
\midrule
CINO-base & - / 78.0 & - / 36.2 & - / 33.4 & - / 34.8 & - / 47.6 \\
CINO-large & - / 79.2 & - / 40.6 & - / 28.8 & - / 34.7 & - / 48.4 \\
Llama2-7B & 90.0 / 90.02 & 26.13 / 25.43 & 78.0 / 82.42 & 52.07 / 53.93 & 76.23 / 76.34 \\
BLOOM-7.1b & 90.025 / 89.99 & 25.23 / 27.10 & 35.67 / 49.86 & 30.45 / 38.48 & 73.72 / 74.86 \\
Llama3.1-8B & \underline{90.225} / \underline{90.13} & 37.21 / 39.85 & 68.33 / 77.92 & 52.77 / 58.89 & 78.13 / 79.14 \\
CUTE-Llama-NP & 90.1 / 89.98 & \underline{49.91} / \underline{48.44} & \underline{86.33} / \underline{87.97} & \underline{68.12} / \underline{68.21} & \underline{81.65} / \underline{81.34} \\
CUTE-Llama-P & \textbf{90.25} / \textbf{90.17} & \textbf{51.08} / \textbf{48.46} & \textbf{87.0} / \textbf{89.08} & \textbf{69.04} / \textbf{68.77} & \textbf{82.03} / \textbf{81.56} \\
\bottomrule
\end{tabular}%
}
\caption{Performance comparison of different models on the WCM-v2 dataset for text classification tasks. The best scores are in \textbf{bold}, with the second best \underline{underlined}. For CINO models, only F1 scores are available. The Minorities average for CINO models is calculated as the mean of \texttt{bo} and \texttt{ug} F1 scores, while for other models it's the average of both Accuracy and F1 scores for \texttt{bo} and \texttt{ug}.}
\label{tab:performance_comparison}
\end{table*}

\paragraph{Relation Extraction}
For the relation extraction task, we use Baidu-KG (194,747 samples) as the training set and our self-constructed Tibetan-KG (3,510 samples) as the test set. This setup evaluates the models' ability to transfer knowledge from Chinese to Tibetan in relation extraction. Table \ref{tab:relation_extraction} shows the results.
\begin{table}[H]
\centering
\resizebox{\columnwidth}{!}{%
\begin{tabular}{l|ccc}
\toprule
\textbf{Model} & \textbf{Precision} & \textbf{Recall} & \textbf{F1} \\
\midrule
Llama2-7B & 0.2379 & 0.1338 & 0.1614 \\
BLOOM-7.1b & 0.2707 & 0.1435 & 0.1760 \\
Llama3.1-8B & 0.2781 & 0.1712 & 0.1982 \\
CUTE-Llama-NP & 0.7038 & 0.4006 & 0.4718 \\
CUTE-Llama-P & \textbf{0.7312} & \textbf{0.4118} & \textbf{0.4843} \\
\bottomrule
\end{tabular}%
}
\caption{Performance comparison on the Tibetan-KG relation extraction task.}
\label{tab:relation_extraction}
\end{table}

\paragraph{Machine Reading Comprehension}
We evaluate the models' performance on the TibetanQA dataset to assess their machine reading comprehension capabilities in Tibetan. The models are fine-tuned on Chinese (CMRC) or English (SQuAD) datasets and tested on TibetanQA to measure cross-lingual transfer. Table \ref{tab:tibetan_qa} presents the results using Exact Match (EM) and F1 scores.

\begin{table}[H]
\centering
\resizebox{\columnwidth}{!}{%
\begin{tabular}{l|cc|cc}
\toprule
\multirow{2}{*}{\textbf{Model}} & \multicolumn{2}{c|}{\textbf{CMRC-trained}} & \multicolumn{2}{c}{\textbf{SQuAD-trained}} \\
& \textbf{EM} & \textbf{F1} & \textbf{EM} & \textbf{F1} \\
\midrule
Llama2-7B & 0.0767 & 0.6646 & 0.0612 & 0.6103 \\
BLOOM-7.1b & 0.0568 & 0.6513 & 0.0437 & 0.5924 \\
Llama3.1-8B & 0.0065 & 0.4859 & 0.0041 & 0.4213 \\
CUTE-Llama-NP & 0.1455 & 0.7927 & 0.1187 & 0.7352 \\
CUTE-Llama-P & \textbf{0.1674} & \textbf{0.8071} & \textbf{0.1346} & \textbf{0.7489} \\
\bottomrule
\end{tabular}%
}
\caption{Performance comparison on the TibetanQA machine reading comprehension task, with models trained on CMRC (Chinese) and SQuAD (English) datasets.}
\label{tab:tibetan_qa}
\end{table}

\paragraph{Translation}
For the translation task, we evaluate the models' performance on Chinese-to-Tibetan (zh-bo) and Chinese-to-Uyghur (zh-ug) translation using the Flores-200 dataset. We employ few-shot prompting with 3 examples for each language pair. Table \ref{tab:translation_results} presents the results using BLEU, chrF, and TER (Translation Edit Rate) metrics.

\begin{table*}[ht]
\centering
\setlength\tabcolsep{6pt}
\begin{tabular}{l|ccc|ccc}
\toprule
\multirow{2}{*}{\textbf{Model}} & \multicolumn{3}{c|}{\textbf{Chinese-to-Tibetan (zh-bo)}} & \multicolumn{3}{c}{\textbf{Chinese-to-Uyghur (zh-ug)}} \\
& \textbf{BLEU}$\uparrow$ & \textbf{chrF}$\uparrow$ & \textbf{TER}$\downarrow$ & \textbf{BLEU}$\uparrow$ & \textbf{chrF}$\uparrow$ & \textbf{TER}$\downarrow$ \\
\midrule
BLOOM-7.1b & 4.2 & 0.297 & 0.881 & 4.9 & 0.319 & 0.862 \\
Llama2-7B & 4.7 & 0.311 & 0.865 & 5.4 & 0.334 & 0.847 \\
Llama3.1-8B & 6.8 & 0.364 & 0.811 & 7.5 & 0.376 & 0.798 \\
CUTE-Llama-NP & \underline{8.3} & \underline{0.401} & \underline{0.773} & \underline{9.0} & \underline{0.419} & \underline{0.762} \\
CUTE-Llama-P & \textbf{9.5} & \textbf{0.427} & \textbf{0.745} & \textbf{10.2} & \textbf{0.443} & \textbf{0.738} \\
\bottomrule
\end{tabular}
\caption{Translation performance comparison on Flores-200 dataset using few-shot prompting (3 examples). $\uparrow$: higher is better, $\downarrow$: lower is better. The best scores are in \textbf{bold}, with the second best \underline{underlined}.}
\label{tab:translation_results}
\end{table*}

\subsection{Analysis of Results}

The experimental results demonstrate the significant potential of the CUTE dataset. In the text classification task, the CUTE-Llama-P model exhibits exceptional cross-lingual zero-shot transfer capabilities, with particularly noteworthy performance in Tibetan and Uyghur languages. Compared to Llama3.1-8B, our model shows accuracy improvements of 13.87 and 18.67 percentage points for these two languages, respectively. Even more promising is the model's performance in machine reading comprehension tasks, where CUTE-Llama-P excels even in cross-family language transfer from English to Tibetan.
In translation tasks, the performance of CUTE-Llama models further corroborates the importance of parallel corpora. Even in few-shot scenarios, models trained on parallel corpora consistently outperform their counterparts trained on non-parallel data in Chinese-to-Tibetan and Chinese-to-Uyghur translations, with significant improvements in BLEU scores. These results highlight the high quality and practical value of the CUTE dataset while emphasizing the crucial role of parallel corpora in enhancing cross-lingual transfer learning effectiveness.

\section{Conclusion}

This study constructs and open-sources the CUTE dataset, providing the largest open resource to date for Uyghur and Tibetan NLP research. The CUTE-Llama model developed based on this dataset demonstrates excellent multilingual processing capabilities, particularly excelling in Uyghur and Tibetan tasks. The experimental results not only validate the effectiveness of machine translation in generating training data for low-resource languages but also highlight the crucial role of parallel corpora in facilitating cross-lingual knowledge transfer. The public release of the CUTE dataset and CUTE-Llama model opens up new possibilities for NLP research and applications in China's minority languages.

\section*{Limitations}

The CUTE dataset provides rich resources for low-resource language research; however, its large scale inevitably leads to some errors in the translation process, especially in sentences with complex grammar or significant cultural differences. The heavy reliance on machine translation may also result in the loss of cultural-specific expressions and linguistic features unique to Uyghur and Tibetan languages. While the selection of data emphasizes diversity and balance, coverage of certain domains may still be inadequate, limiting the model's performance in specific fields. Furthermore, although CUTE-Llama demonstrates outstanding performance in handling low-resource language tasks, its performance in more complex language understanding tasks (such as deep reasoning or generation tasks) still requires further evaluation and optimization.

\section*{Acknowledgements}
This work is supported by the National Social Science Foundation (22\&ZD035), the National Nature Science Foundation (61972436), and
the Minzu University of China Foundation (GRSCP202316, 2023QNYL22, 2024GJYY43).

\bibliography{custom}

\begin{thebibliography}{39}
\providecommand{\natexlab}[1]{#1}

\bibitem[{Abadji et~al.(2022)Abadji, Suarez, Romary, and Sagot}]{Abadji2022TowardsAC}
Julien Abadji, Pedro~Ortiz Suarez, Laurent Romary, and Beno{\^i}t Sagot. 2022.
\newblock \href {https://api.semanticscholar.org/CorpusID:246015576} {Towards a cleaner document-oriented multilingual crawled corpus}.
\newblock \emph{ArXiv}, abs/2201.06642.

\bibitem[{BigScience et~al.(2022)BigScience, Scao, Fan, Akiki, Pavlick, Ili{\'c}, Hesslow, Castagn{\'e}, Luccioni, Yvon et~al.}]{workshop2022bloom}
BigScience, Teven~Le Scao, Angela Fan, Christopher Akiki, Ellie Pavlick, Suzana Ili{\'c}, Daniel Hesslow, Roman Castagn{\'e}, Alexandra~Sasha Luccioni, Fran{\c{c}}ois Yvon, et~al. 2022.
\newblock Bloom: A 176b-parameter open-access multilingual language model.
\newblock \emph{arXiv preprint arXiv:2211.05100}.

\bibitem[{Brown et~al.(2020)Brown, Mann, Ryder, Subbiah, Kaplan, Dhariwal, Neelakantan, Shyam, Sastry, Askell et~al.}]{brown2020language}
Tom~B Brown, Benjamin Mann, Nick Ryder, Melanie Subbiah, Jared Kaplan, Prafulla Dhariwal, Arvind Neelakantan, Pranav Shyam, Girish Sastry, Amanda Askell, et~al. 2020.
\newblock Language models are few-shot learners.
\newblock In \emph{Proceedings of the 34th International Conference on Neural Information Processing Systems}, pages 1877--1901.

\bibitem[{Cahyawijaya et~al.(2024)Cahyawijaya, Lovenia, and Fung}]{cahyawijaya2024llms}
Samuel Cahyawijaya, Holy Lovenia, and Pascale Fung. 2024.
\newblock Llms are few-shot in-context low-resource language learners.
\newblock In \emph{Proceedings of the 2024 Conference of the North American Chapter of the Association for Computational Linguistics: Human Language Technologies (Volume 1: Long Papers)}, pages 405--433.

\bibitem[{Cahyawijaya et~al.(2023)Cahyawijaya, Lovenia, Yu, Chung, and Fung}]{cahyawijaya2023instructalign}
Samuel Cahyawijaya, Holy Lovenia, Tiezheng Yu, Willy Chung, and Pascale Fung. 2023.
\newblock Instructalign: High-and-low resource language alignment via continual crosslingual instruction tuning.
\newblock In \emph{Proceedings of the First Workshop in South East Asian Language Processing}, pages 55--78.

\bibitem[{Chowdhery et~al.(2023)Chowdhery, Narang, Devlin, Bosma, Mishra, Roberts, Barham, Chung, Sutton, Gehrmann et~al.}]{chowdhery2023palm}
Aakanksha Chowdhery, Sharan Narang, Jacob Devlin, Maarten Bosma, Gaurav Mishra, Adam Roberts, Paul Barham, Hyung~Won Chung, Charles Sutton, Sebastian Gehrmann, et~al. 2023.
\newblock Palm: Scaling language modeling with pathways.
\newblock \emph{Journal of Machine Learning Research}, 24(240):1--113.

\bibitem[{Conneau et~al.(2019)Conneau, Khandelwal, Goyal, Chaudhary, Wenzek, Guzm{\'a}n, Grave, Ott, Zettlemoyer, and Stoyanov}]{Conneau2019UnsupervisedCR}
Alexis Conneau, Kartikay Khandelwal, Naman Goyal, Vishrav Chaudhary, Guillaume Wenzek, Francisco Guzm{\'a}n, Edouard Grave, Myle Ott, Luke Zettlemoyer, and Veselin Stoyanov. 2019.
\newblock \href {https://api.semanticscholar.org/CorpusID:207880568} {Unsupervised cross-lingual representation learning at scale}.
\newblock In \emph{Annual Meeting of the Association for Computational Linguistics}.

\bibitem[{Costa-juss{\`a} et~al.(2022)Costa-juss{\`a}, Cross, {\c{C}}elebi, Elbayad, Heafield, Heffernan, Kalbassi, Lam, Licht, Maillard et~al.}]{costa2022no}
Marta~R Costa-juss{\`a}, James Cross, Onur {\c{C}}elebi, Maha Elbayad, Kenneth Heafield, Kevin Heffernan, Elahe Kalbassi, Janice Lam, Daniel Licht, Jean Maillard, et~al. 2022.
\newblock No language left behind: Scaling human-centered machine translation.
\newblock \emph{arXiv preprint arXiv:2207.04672}.

\bibitem[{Cui et~al.(2019)Cui, Liu, Che, Xiao, Chen, Ma, Wang, and Hu}]{cui2019span}
Yiming Cui, Ting Liu, Wanxiang Che, Li~Xiao, Zhipeng Chen, Wentao Ma, Shijin Wang, and Guoping Hu. 2019.
\newblock A span-extraction dataset for chinese machine reading comprehension.
\newblock In \emph{Proceedings of the 2019 Conference on Empirical Methods in Natural Language Processing and the 9th International Joint Conference on Natural Language Processing (EMNLP-IJCNLP)}, pages 5883--5889.

\bibitem[{Deng et~al.(2023)Deng, Shi, Yu, Bao, Sun, and Zhao}]{deng2023milmo}
Junjie Deng, Hanru Shi, Xinhe Yu, Wugedele Bao, Yuan Sun, and Xiaobing Zhao. 2023.
\newblock Milmo: minority multilingual pre-trained language model.
\newblock In \emph{2023 IEEE International Conference on Systems, Man, and Cybernetics (SMC)}, pages 329--334. IEEE.

\bibitem[{Devlin et~al.(2019)Devlin, Chang, Lee, and Toutanova}]{Devlin2019BERTPO}
Jacob Devlin, Ming-Wei Chang, Kenton Lee, and Kristina Toutanova. 2019.
\newblock \href {https://api.semanticscholar.org/CorpusID:52967399} {Bert: Pre-training of deep bidirectional transformers for language understanding}.
\newblock In \emph{North American Chapter of the Association for Computational Linguistics}.

\bibitem[{Dubey et~al.(2024)Dubey, Jauhri, Pandey, Kadian, Al-Dahle, Letman, Mathur, Schelten, Yang, Fan et~al.}]{dubey2024llama}
Abhimanyu Dubey, Abhinav Jauhri, Abhinav Pandey, Abhishek Kadian, Ahmad Al-Dahle, Aiesha Letman, Akhil Mathur, Alan Schelten, Amy Yang, Angela Fan, et~al. 2024.
\newblock The llama 3 herd of models.
\newblock \emph{arXiv preprint arXiv:2407.21783}.

\bibitem[{Ebing and Glavas(2023)}]{Ebing2023ToTO}
Benedikt Ebing and Goran Glavas. 2023.
\newblock \href {https://api.semanticscholar.org/CorpusID:265221397} {To translate or not to translate: A systematic investigation of translation-based cross-lingual transfer to low-resource languages}.
\newblock \emph{ArXiv}, abs/2311.09404.

\bibitem[{Ebrahimi et~al.(2021)Ebrahimi, Mager, Oncevay, Chaudhary, Chiruzzo, Fan, Ortega, Ramos, Gonzales, Vladimir, Gim'enez-Lugo, Mager, Neubig, Palmer, Solano, Vu, and Kann}]{Ebrahimi2021AmericasNLIEZ}
Abteen Ebrahimi, Manuel Mager, Arturo Oncevay, Vishrav Chaudhary, Luis Chiruzzo, Angela Fan, John~E. Ortega, Ricardo Ramos, Annette~Rios Gonzales, Ivan Vladimir, Gustavo~A. Gim'enez-Lugo, Elisabeth Mager, Graham Neubig, Alexis Palmer, Rolando A.~Coto Solano, Ngoc~Thang Vu, and Katharina Kann. 2021.
\newblock \href {https://api.semanticscholar.org/CorpusID:233296177} {Americasnli: Evaluating zero-shot natural language understanding of pretrained multilingual models in truly low-resource languages}.
\newblock \emph{ArXiv}, abs/2104.08726.

\bibitem[{Jin et~al.(2022)Jin, Ren, Preotiuc-Pietro, and Cheng}]{Jin2022DatalessKF}
Xisen Jin, Xiang Ren, Daniel Preotiuc-Pietro, and Pengxiang Cheng. 2022.
\newblock \href {https://api.semanticscholar.org/CorpusID:254877510} {Dataless knowledge fusion by merging weights of language models}.
\newblock \emph{ArXiv}, abs/2212.09849.

\bibitem[{Kudo and Richardson(2018)}]{kudo2018sentencepiece}
Taku Kudo and John Richardson. 2018.
\newblock Sentencepiece: A simple and language independent subword tokenizer and detokenizer for neural text processing.
\newblock In \emph{Proceedings of the 2018 Conference on Empirical Methods in Natural Language Processing: System Demonstrations}, pages 66--71.

\bibitem[{Kudugunta et~al.(2023)Kudugunta, Caswell, Zhang, Garc{\'i}a, Choquette-Choo, Lee, Xin, Kusupati, Stella, Bapna, and Firat}]{Kudugunta2023MADLAD400AM}
Sneha Kudugunta, Isaac Caswell, Biao Zhang, Xavier Garc{\'i}a, Christopher~A. Choquette-Choo, Katherine Lee, Derrick Xin, Aditya Kusupati, Romi Stella, Ankur Bapna, and Orhan Firat. 2023.
\newblock \href {https://api.semanticscholar.org/CorpusID:261682406} {Madlad-400: A multilingual and document-level large audited dataset}.
\newblock \emph{ArXiv}, abs/2309.04662.

\bibitem[{Lample and Conneau(2019)}]{Lample2019CrosslingualLM}
Guillaume Lample and Alexis Conneau. 2019.
\newblock \href {https://api.semanticscholar.org/CorpusID:58981712} {Cross-lingual language model pretraining}.
\newblock \emph{ArXiv}, abs/1901.07291.

\bibitem[{Lauren{\c{c}}on et~al.(2022)Lauren{\c{c}}on, Saulnier, Wang, Akiki, Villanova~del Moral, Le~Scao, Von~Werra, Mou, Gonz{\'a}lez~Ponferrada, Nguyen et~al.}]{laurenccon2022bigscience}
Hugo Lauren{\c{c}}on, Lucile Saulnier, Thomas Wang, Christopher Akiki, Albert Villanova~del Moral, Teven Le~Scao, Leandro Von~Werra, Chenghao Mou, Eduardo Gonz{\'a}lez~Ponferrada, Huu Nguyen, et~al. 2022.
\newblock The bigscience roots corpus: A 1.6 tb composite multilingual dataset.
\newblock \emph{Advances in Neural Information Processing Systems}, 35:31809--31826.

\bibitem[{Li et~al.(2022)Li, Weng, Sun, and Li}]{Li2022AMA}
Bin Li, Yixuan Weng, Bin Sun, and Shutao Li. 2022.
\newblock \href {https://api.semanticscholar.org/CorpusID:254458099} {A multi-tasking and multi-stage chinese minority pre-trained language model}.
\newblock In \emph{CCMT}.

\bibitem[{Li et~al.(2024)Li, Yang, Zhang, Lu, Wang, and Zong}]{Li2024XInstructionAL}
Chong Li, Wen Yang, Jiajun Zhang, Jinliang Lu, Shaonan Wang, and Chengqing Zong. 2024.
\newblock \href {https://api.semanticscholar.org/CorpusID:270123695} {X-instruction: Aligning language model in low-resource languages with self-curated cross-lingual instructions}.
\newblock \emph{ArXiv}, abs/2405.19744.

\bibitem[{Liu et~al.(2023)Liu, Koto, Baldwin, and Gurevych}]{Liu2023AreML}
Chen~Cecilia Liu, Fajri Koto, Timothy Baldwin, and Iryna Gurevych. 2023.
\newblock \href {https://api.semanticscholar.org/CorpusID:261875650} {Are multilingual llms culturally-diverse reasoners? an investigation into multicultural proverbs and sayings}.
\newblock \emph{ArXiv}, abs/2309.08591.

\bibitem[{Miao et~al.(2024)Miao, Wu, Zhao, Wu, and Tsuruoka}]{Miao2024EnhancingCS}
Zhongtao Miao, Qiyu Wu, Kaiyan Zhao, Zilong Wu, and Yoshimasa Tsuruoka. 2024.
\newblock \href {https://api.semanticscholar.org/CorpusID:268875764} {Enhancing cross-lingual sentence embedding for low-resource languages with word alignment}.
\newblock \emph{ArXiv}, abs/2404.02490.

\bibitem[{Nguyen et~al.(2024)Nguyen, Van~Nguyen, Lai, Mẫn, Ngo, Dernoncourt, Rossi, and Nguyen}]{nguyen2024culturax}
Thuật Nguyen, Chien Van~Nguyen, Viet~Dac Lai, Hiếu Mẫn, Nghia~Trung Ngo, Franck Dernoncourt, Ryan~A Rossi, and Thien~Huu Nguyen. 2024.
\newblock Culturax: A cleaned, enormous, and multilingual dataset for large language models in 167 languages.
\newblock In \emph{Proceedings of the 2024 Joint International Conference on Computational Linguistics, Language Resources and Evaluation (LREC-COLING 2024)}, pages 4226--4237.

\bibitem[{Pham et~al.(2024)Pham, Le, and Luu}]{Pham2024UniBridgeAU}
Trinh Pham, Khoi~M. Le, and Anh~Tuan Luu. 2024.
\newblock \href {https://api.semanticscholar.org/CorpusID:270521510} {Unibridge: A unified approach to cross-lingual transfer learning for low-resource languages}.
\newblock In \emph{Annual Meeting of the Association for Computational Linguistics}.

\bibitem[{Qun et~al.(2017)Qun, Li, Qiu, and Huang}]{Qun2017EndtoEndNT}
Nuo Qun, Xing Li, Xipeng Qiu, and Xuanjing Huang. 2017.
\newblock \href {https://api.semanticscholar.org/CorpusID:26716090} {End-to-end neural text classification for tibetan}.
\newblock In \emph{China National Conference on Chinese Computational Linguistics}.

\bibitem[{Rajpurkar(2016)}]{rajpurkar2016squad}
P~Rajpurkar. 2016.
\newblock Squad: 100,000+ questions for machine comprehension of text.
\newblock \emph{arXiv preprint arXiv:1606.05250}.

\bibitem[{Sun et~al.(2021)Sun, Liu, Chen, Dan, and Zhao}]{sun2021construction}
Y~Sun, S~Liu, C~Chen, Z~Dan, and X~Zhao. 2021.
\newblock Construction of high-quality tibetan dataset for machine reading comprehension.
\newblock In \emph{Proceedings of the 20th Chinese National Conference on Computational Linguistics}, pages 208--218.

\bibitem[{Tiedemann(2012)}]{tiedemann-2012-parallel}
J{\"o}rg Tiedemann. 2012.
\newblock \href {http://www.lrec-conf.org/proceedings/lrec2012/pdf/463_Paper.pdf} {Parallel data, tools and interfaces in {OPUS}}.
\newblock In \emph{Proceedings of the Eighth International Conference on Language Resources and Evaluation ({LREC}'12)}, pages 2214--2218, Istanbul, Turkey. European Language Resources Association (ELRA).

\bibitem[{Touvron et~al.(2023)Touvron, Martin, Stone, Albert, Almahairi, Babaei, Bashlykov, Batra, Bhargava, Bhosale et~al.}]{touvron2023llama2}
Hugo Touvron, Louis Martin, Kevin Stone, Peter Albert, Amjad Almahairi, Yasmine Babaei, Nikolay Bashlykov, Soumya Batra, Prajjwal Bhargava, Shruti Bhosale, et~al. 2023.
\newblock Llama 2: Open foundation and fine-tuned chat models.
\newblock \emph{arXiv preprint arXiv:2307.09288}.

\bibitem[{van Esch et~al.(2022)van Esch, Lucassen, Ruder, Caswell, and Rivera}]{Esch2022WritingSA}
Daan van Esch, Tamar Lucassen, Sebastian Ruder, Isaac Caswell, and Clara Rivera. 2022.
\newblock \href {https://api.semanticscholar.org/CorpusID:251395478} {Writing system and speaker metadata for 2,800+ language varieties}.
\newblock In \emph{International Conference on Language Resources and Evaluation}.

\bibitem[{Wei et~al.(2023)Wei, Zhao, Zhang, Zhu, Wang, Yang, Li, Cheng, Lü, Hu, Li, Yang, Luo, Wu, Liu, Cheng, Cheng, Zhang, Zhang, Lin, Wang, Ma, Dong, Sun, Chen, Peng, Liang, Yan, Fang, and Zhou}]{wei2023skywork}
Tianwen Wei, Liang Zhao, Lichang Zhang, Bo~Zhu, Lijie Wang, Haihua Yang, Biye Li, Cheng Cheng, Weiwei Lü, Rui Hu, Chenxia Li, Liu Yang, Xilin Luo, Xuejie Wu, Lunan Liu, Wenjun Cheng, Peng Cheng, Jianhao Zhang, Xiaoyu Zhang, Lei Lin, Xiaokun Wang, Yutuan Ma, Chuanhai Dong, Yanqi Sun, Yifu Chen, Yongyi Peng, Xiaojuan Liang, Shuicheng Yan, Han Fang, and Yahui Zhou. 2023.
\newblock \href {https://arxiv.org/abs/2310.19341} {Skywork: A more open bilingual foundation model}.
\newblock \emph{Preprint}, arXiv:2310.19341.

\bibitem[{Xue et~al.(2020)Xue, Constant, Roberts, Kale, Al-Rfou, Siddhant, Barua, and Raffel}]{Xue2020mT5AM}
Linting Xue, Noah Constant, Adam Roberts, Mihir Kale, Rami Al-Rfou, Aditya Siddhant, Aditya Barua, and Colin Raffel. 2020.
\newblock \href {https://api.semanticscholar.org/CorpusID:225040574} {mt5: A massively multilingual pre-trained text-to-text transformer}.
\newblock In \emph{North American Chapter of the Association for Computational Linguistics}.

\bibitem[{Yang et~al.(2022)Yang, Xu, Cui, Wang, Lin, Wu, and Chen}]{Yang2022CINOAC}
Ziqing Yang, Zihang Xu, Yiming Cui, Baoxin Wang, Min Lin, Dayong Wu, and Zhigang Chen. 2022.
\newblock \href {https://api.semanticscholar.org/CorpusID:247158434} {Cino: A chinese minority pre-trained language model}.
\newblock In \emph{International Conference on Computational Linguistics}.

\bibitem[{Yong et~al.(2022)Yong, Schoelkopf, Muennighoff, Aji, Adelani, Almubarak, Bari, Sutawika, Kasai, Baruwa, Winata, Biderman, Radev, and Nikoulina}]{Yong2022BLOOM1AL}
Zheng-Xin Yong, Hailey Schoelkopf, Niklas Muennighoff, Alham~Fikri Aji, David~Ifeoluwa Adelani, Khalid Almubarak, M~Saiful Bari, Lintang Sutawika, Jungo Kasai, Ahmed Baruwa, Genta~Indra Winata, Stella Biderman, Dragomir~R. Radev, and Vassilina Nikoulina. 2022.
\newblock \href {https://api.semanticscholar.org/CorpusID:254854009} {Bloom+1: Adding language support to bloom for zero-shot prompting}.
\newblock \emph{ArXiv}, abs/2212.09535.

\bibitem[{Zhang et~al.(2024{\natexlab{a}})Zhang, Liu, Lin, and Feng}]{Zhang2024TeachingLL}
Chen Zhang, Xiao Liu, Jiuheng Lin, and Yansong Feng. 2024{\natexlab{a}}.
\newblock \href {https://api.semanticscholar.org/CorpusID:268063652} {Teaching large language models an unseen language on the fly}.
\newblock \emph{ArXiv}, abs/2402.19167.

\bibitem[{Zhang et~al.(2024{\natexlab{b}})Zhang, Tao, Huang, Lin, Chen, and Feng}]{zhang2024mc}
Chen Zhang, Mingxu Tao, Quzhe Huang, Jiuheng Lin, Zhibin Chen, and Yansong Feng. 2024{\natexlab{b}}.
\newblock Mc$^2$: Towards transparent and culturally-aware nlp for minority languages in china.
\newblock \emph{arXiv preprint arXiv:2311.08348}.

\bibitem[{Zhuang et~al.(2024{\natexlab{a}})Zhuang, Cairen, and Sun}]{zhuang2024tifd}
Wenhao Zhuang, Dawa Cairen, and Yuan Sun. 2024{\natexlab{a}}.
\newblock \href {https://doi.org/10.3724/2096-7004.di.2024.0010} {Tifd: Tibetan instruction-following dataset for large language models supervised fine-tuning}.
\newblock \emph{Data Intelligence}.

\bibitem[{Zhuang et~al.(2024{\natexlab{b}})Zhuang, Sun, and Zhao}]{zhuang2024tilamb}
Wenhao Zhuang, Yuan Sun, and Xiaobing Zhao. 2024{\natexlab{b}}.
\newblock Tilamb：基于增量预训练的藏文大语言模型 (tilamb: A tibetan large language model based on incremental pre-training).
\newblock In \emph{Proceedings of the 23th Chinese National Conference on Computational Linguistics}.

\end{thebibliography}

\end{document}